**RESEARCH ARTICLE**

WILEY

# Group-level brain decoding with deep learning

Richard Csaky[1,2,3] | Mats W. J. van Es[1,2] | Oiwi Parker Jones[2,4,5] | Mark Woolrich[1,2]

[1]Oxford Centre for Human Brain Activity, Department of Psychiatry, University of Oxford, Oxford, UK

[2]Wellcome Centre for Integrative Neuroimaging, Oxford, UK

[3]Christ Church, Oxford, UK

[4]Jesus College, Oxford, UK

[5]Department of Engineering Science, University of Oxford, Oxford, UK

**Correspondence**
Richard Csaky, Oxford Centre for Human Brain Activity, Department of Psychiatry, University of Oxford, OX3 7JX, Oxford, UK.
Email: richard.csaky@psych.ox.ac.uk

**Funding information**
Wellcome Centre Integrative Neuroimaging Studentship; Wellcome Trust, Grant/Award Numbers: 203139/Z/16/Z, 215573/Z/19/Z, 106183/Z/14/Z; UK MRC, Grant/Award Number: MR/X00757X/1; Dementia Platform UK, Grant/Award Number: RG94383/RG89702; EU-project euSNN, Grant/Award Number: MSCA-ITN H2020-860563

**Abstract**

Decoding brain imaging data are gaining popularity, with applications in brain-computer interfaces and the study of neural representations. Decoding is typically subject-specific and does not generalise well over subjects, due to high amounts of between subject variability. Techniques that overcome this will not only provide richer neuroscientific insights but also make it possible for group-level models to outperform subject-specific models. Here, we propose a method that uses subject embedding, analogous to word embedding in natural language processing, to learn and exploit the structure in between-subject variability as part of a decoding model, our adaptation of the WaveNet architecture for classification. We apply this to magnetoencephalography data, where 15 subjects viewed 118 different images, with 30 examples per image; to classify images using the entire 1 s window following image presentation. We show that the combination of deep learning and subject embedding is crucial to closing the performance gap between subject- and group-level decoding models. Importantly, group models outperform subject models on low-accuracy subjects (although slightly impair high-accuracy subjects) and can be helpful for initialising subject models. While we have not generally found group-level models to perform better than subject-level models, the performance of group modelling is expected to be even higher with bigger datasets. In order to provide physiological interpretation at the group level, we make use of permutation feature importance. This provides insights into the spatiotemporal and spectral information encoded in the models. All code is available on GitHub (https://github.com/ricsinaruto/MEG-group-decode).

## 1 | INTRODUCTION

In recent years, decoding has gained in popularity in neuroimaging (Du et al., 2023; Kay et al., 2008; Lin et al., 2022; Takagi & Nishimoto, 2023), specifically decoding external variables (e.g., stimulus category) from internal states (i.e., brain activity). Such analyses can be useful for brain-computer interface (BCI) applications (Willett et al., 2021) or to gain neuroscientific insights (Cichy et al., 2016; Guggenmos et al., 2018; Kay et al., 2008). Applications of decoding to brain recordings typically fit separate (often linear) models per dataset, per subject (Csaky et al., 2023; Dash et al., 2020a; Guggenmos et al., 2018). This has the benefit that the decoding is tuned to the dataset/subject, but has the drawback that it is unable to leverage knowledge that could be transferred across datasets/subjects. This is especially desirable for the field of neuroimaging because gathering more data are expensive and often impossible (e.g., in clinical populations). More practical drawbacks of subject-specific (subject-level) models include increased computational load, a higher chance of overfitting and the inability to adapt to new subjects. We aim to leverage data from multiple subjects and train a shared model that can generalise across subjects (group-level). A conceptual visualisation of subject-level and group-level models is given in Figure 1.

Due to high temporal resolution and relatively good spatial resolution, magnetoencephalography (MEG) is an excellent method for studying the fast dynamics of brain activity. MEG is highly suitable for decoding analyses (Du et al., 2019), which are mostly done using subject-level models. This is because there are high amounts of between-subject variability in neuroimaging data. An alternative approach would be to train and use the same decoding model across multiple subjects (Li et al., 2021; Olivetti et al., 2014). We will refer to an approach that does this, while not explicitly modelling any of the between-subject variability, as 'naïve group modelling'. Such naïve approaches, effectively pretend that all data comes from the same subject (see Figure 1b), but due to high amounts of between-subject variability typically perform very badly (Li et al., 2021; Olivetti et al., 2014; Saha & Baumert, 2020). The work in this paper is motivated by a need to improve on these methods. If group modelling could be advanced to account for the high amounts of between-subject variability, then this would allow relevant information to be pooled across subjects, resulting in two key benefits. First, we would be able to obtain neuroscientific insights from the decoding models directly at the group level instead of pooling over subjects. Second, with appropriately large multi-subject datasets, group models would be able to outperform subject-level models.

Our main aim is to improve subject-level models by using a single group decoding model that generalises across (and within) subjects. We refer to this as *across-subject* decoding, in which models are trained on part of the data from all subjects and then tested on left-out data from all subjects. This is motivated by the fact that group-level models that perform well in this manner can be useful for gaining neuroscientific insights that are relevant at the group level, as we will show in Sections 3.4 and 3.5. An alternative approach, *Leave-one-subject-out (LOSO) analysis* is also presented in Section 3.3. In LOSO analyses, group-level models are trained on data from multiple subjects and tested on a new, unseen subject (Zubarev et al., 2019), which can be especially useful in zero-shot BCI applications.

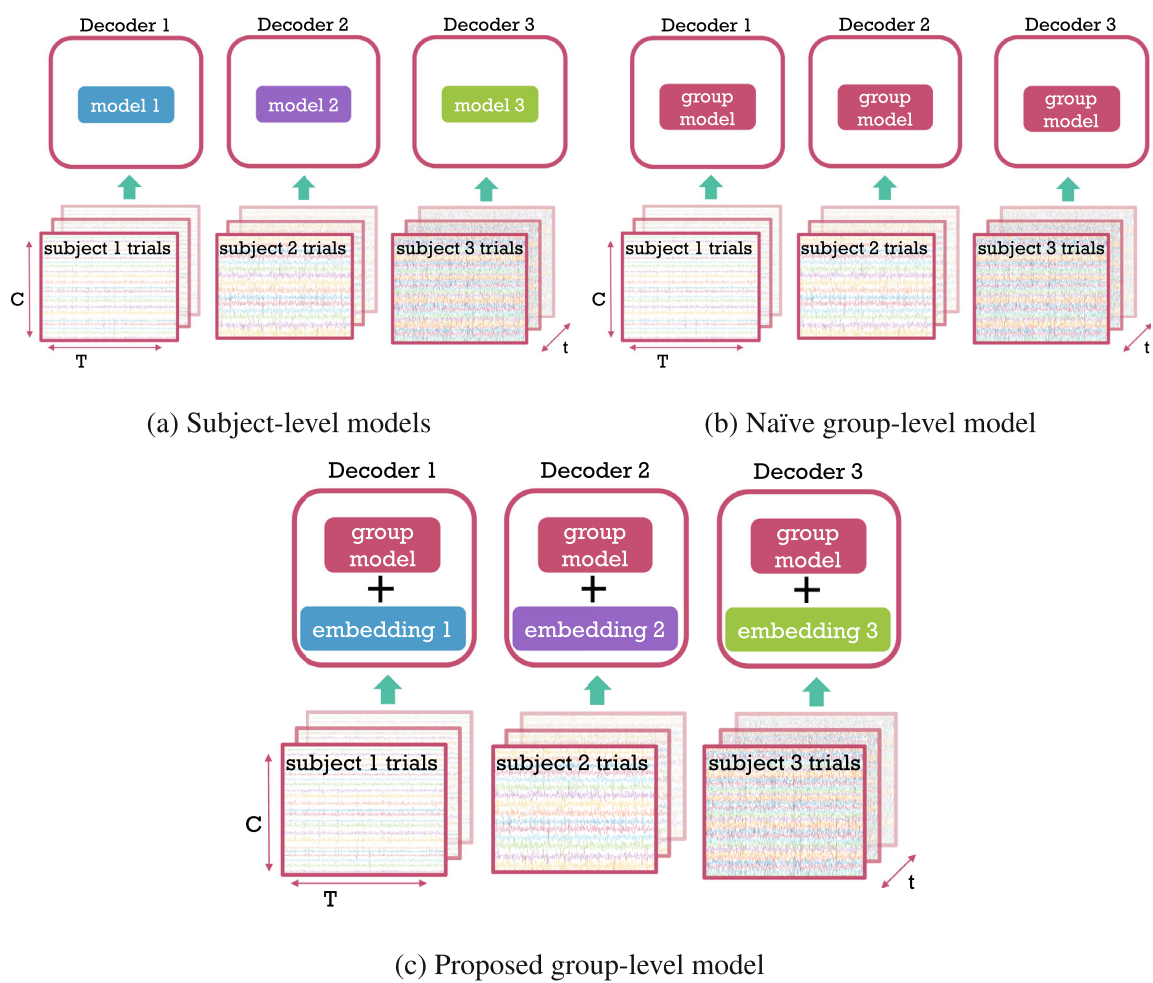

(a) Subject-level models

(b) Naïve group-level model

(c) Proposed group-level model

**FIGURE 1** Comparison of subject-level (a), naive group-level (b), the proposed group-level (c) modelling. (a) A separate model is trained on the trials (examples) of each subject. (b) A single, shared model is trained on the trials of all subjects without capturing between-subject variability. (c) A single, shared model is trained on the trials of all subjects with an additional embedding component that is subject-specific. Each trial is **C** × **T** (channels × time points) dimensional. Each of the **s** subjects has **t** trials.

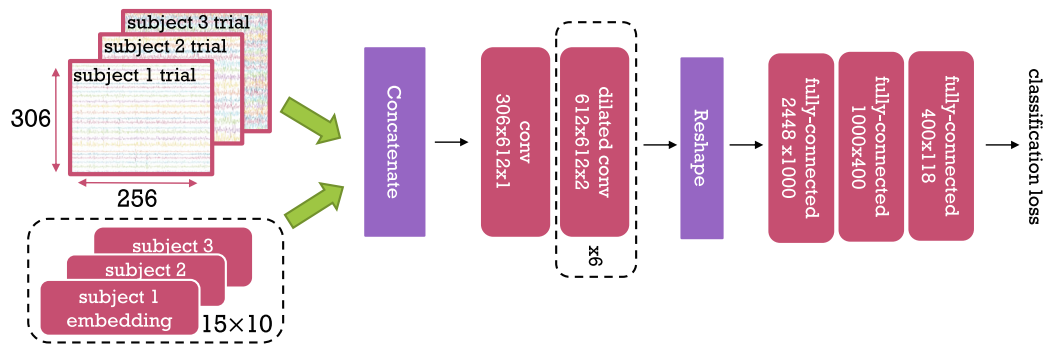

**FIGURE 2** Group-level WaveNet Classifier with subject embeddings. Dashed boxes represent parts of the model which differ between subject-level and group-level versions of our architecture. Red boxes represent learnable parameters. For convolutional layers, the numbers represent *input channels × output channels × kernel size*. For fully-connected layers, the numbers represent *input neurons × output neurons*. The embedding layer dimensionality is given as **s** × **E** (15 × 10), where **s** is number of subjects, and **E** is the embedding size. Embeddings are concatenated with input trials to provide information about which trial is coming from which subject. The classification loss is cross-entropy.

Here, we propose a general architecture capable of jointly decoding multiple subjects with the help of subject embeddings (Figures 1c and 2). Note that we do this in the context of full-epoch decoding, as it has been recently shown that full-epoch models perform better on our dataset of choice than sliding-window decoding (Csaky et al., 2023). Using a 15-subject MEG dataset with a visual task (Cichy et al., 2016), we make the following contributions. First, we introduce a group-level model with subject embeddings, substantially improving over naive group modelling and showing the potential improvements in decoding that can be provided over subject-specific decoding models. Second, we provide insight into how non-linearity and subject embedding helps group modelling. Third, we show that we can gain neuroscientific insights from the deep learning-based decoding model, using permutation feature importance (Altmann et al., 2010) to reveal how meaningful spatiotemporal and spectral information is encoded. While PFI has been used before to provide insights at the group-level by pooling over linear subject-level models (Csaky et al., 2023), here we show that it works similarly well on non-linear group-level models.

## 2 | METHODS

### 2.1 | Data

In this work, a task-MEG dataset is used where 15 subjects view 118 different images, with each image viewed 30 times (Cichy et al., 2016). The raw epoched data are publicly available,[1] however, we obtained the continuous raw MEG data directly from the authors to be able to run our preprocessing pipeline using MNE-Python (Gramfort et al., 2013). Our preprocessed data can be reproduced by running the same pipeline on the continuous raw data. However, one may also use the publicly available epoched data to closely match our results.

Raw data are bandpass filtered between 0.1 and 125 Hz and line noise is removed with notch filters. Whitening is used to remove covariance between channels for subject-level models. Removal of cross-channel covariance (whitening), or in other words multivariate noise normalisation has been previously found to improve the performance of linear decoding models (Guggenmos et al., 2018). The whitening is simply done by performing a PCA projection over the channels keeping all components. For group-level models no whitening is performed, instead, each channel is individually standardised by removing the mean and dividing by the variance. The reason for not using whitening in the case of group-level models is that it would destroy the alignment of the channels between subjects, as each PCA decomposition projects into a different space. Alternatively, we can run PCA at the group-level on the data concatenated over subjects, however, we did not see an improvement in performance when doing this (data not shown).

After whitening, we downsample to 250 Hz and 1.024-second epochs are extracted, starting 100 ms before stimulus presentation. This resulted in 306 × 256-dimensional trials (channels × time points) from the 306 MEG sensors. We do multiclass decoding, predicting a separate probability for each of the 118 classes (images). For a summary of the epoched data see Table 1.

**TABLE 1** Dimensions of the epoched dataset.

| Number of subjects | Number of classes | Number of samples per class | Dimensions of one sample |
|---|---|---|---|
| 15 | 118 | 30 | 306 channels × 256 timesteps |

---

[1] http://userpage.fu-berlin.de/rmcichy/fusion_project_page/main.html

## 2.2 | Models

Our choice of core decoding model was based on a desire to assess the extent to which group decoding models might allow for the use of more complex, non-linear networks when compared to subject-specific decoding. In addition, we did not aim to design a new kind of architecture for decoding MEG data, but rather build our model based on CNN-based architectures that have already been proven to be effective on time series data. As such, we used a decoding model based on WaveNet for classification, which has been used successfully in the audio domain (van den Oord et al., 2016; Zhang et al., 2020), and which we refer to as the Wavenet Classifier. The dilated convolutions in WaveNet are effective for modelling time series data, as successive layers extract complementary frequency content of the input (Borovykh et al., 2018). While CNN-based architectures have been used successfully on M/EEG data (Lawhern et al., 2018), there is no prior work specifically applying Wavenet to neural decoding. To be clear when we refer to *model* in this section we mean the general (untrained) architecture and not a trained model on some dataset. For all training instances in this paper, we used a randomly initialised model instead of using the pretrained weights from the audio-WaveNet.

Our Wavenet Classifier model consists of 2 parts: the (temporal) convolutional block, intended to act as a feature extractor; and the fully-connected block, which is designed for classification (Figure 2). The convolutional block uses a stack of 1D dilated convolutional layers, which include dropout and the inverse hyperbolic sine activation function. For subject-level modelling, we use 3 convolutional layers. For group-level modelling, we use 6 convolutional layers. We arrived at these numbers empirically by training both 3-layer and 6-layer subject-level and group-level models and selecting the best model version in each case, thus providing a fair comparison between subject- and group-level settings. Since the dilation factor is doubled in successive layers, the receptive field of the convolutional block is $2^{num\_layers}$. Given there is no pooling and a convolution stride of 1, the output of each layer preserves the temporal dimensionality.[2] At the end of the convolutional block, we downsample temporally by the size of the receptive field. For example, in the model with 6 convolutional layers, this means that the initial input of size 256 is downsampled by a factor of 64, resulting in 4 values per channel. This downsampling is due to the fact that a receptive field of 64 means that the convolutional part of the model 'sees' 64 time point segments from the 256-time point input trial. Next, this downsampled output is flattened and fed into a fully-connected block. The final output is a logit vector corresponding to the 118 classes. The model is trained with the cross-entropy loss for classification, which includes a softmax function that maps the logit vector to a probability distribution over classes.

We assess two versions of each model, one with a Wavenet Classifier that is linear and one that is non-linear. This allows us to see how non-linearities (a bedrock of deep learning) interact with group modelling. The linear versions simply correspond to Wavenet Classifier where the activation function is set to be the identity function.

Finally, we divide the group-level modelling into two approaches. First, we have a naive group model, which is our standard 6-layer Wavenet Classifier. Second, we have our proposed group model, which improves on the naive group model through the inclusion of subject embeddings. A high-level mathematical description of subject-level (Equation (1)), naïve group-level (Equation 2) and the embedding-aided group-level (Equation (3)) models is given below (corresponding to the 3 panels in Figure 2).

$$\forall s \in S : t_s = f_s(y_s) \quad (1)$$

$$\forall s \in S : t_s = f_g(y_s) \quad (2)$$

$$\forall s \in S : t_s = f_g(y_s, e_s) \quad (3)$$

where $s$ denotes a single subject and $S$ is the set of all subjects. $t_s$ and $y_s$ are the target variables and input trials of subject $s$, $f_s$ is the subject-specific model and $f_g$ is the shared group-level model across subjects. $e_s$ is the subject-specific learned embedding.

Subject embeddings are introduced as a way of dealing with between-subject variability, similarly to Chehab et al. (2022). Like word embeddings in NLP, each subject has a corresponding dense vector (Mikolov et al., 2013). This same vector is concatenated with the channel dimension of the input trial across all time points (in each trial). This operation is given in programming notation below.

$$x_s = \text{concatenate}((y_s, e_s), \dim = 0) \quad (4)$$

where $y_s \in R^{C \times T}$ is the input trial consisting of $C$ channels and $T$ time points, $e_s \in R^{E \times 1}$ is the subject embedding of size $E$ and $x_s \in R^{(C+E) \times T}$ is the input that gets fed into the model ($f_g$). Embedding size was set to 10 a priori, and the effect of different values is explored in Section 3.2. Subject embeddings are learnt together with other model weights using backpropagation. We reasoned that an embedding-aided model can learn general features across subjects, with the capability of adapting its internal representations for each subject.

## 2.3 | Model analysis

In this section, we describe several approaches to uncovering the information encoded in the WaveNet Classifier. In *Kernel FIR Analysis*, we investigate the frequency characteristics of the convolutional kernels. Random noise is fed into a trained model, and the power spectral density of the output of specific kernels is computed to assess their finite impulse response (FIR) properties.

Permutation feature importance (PFI) is a powerful method to assess which features contribute the most to model performance (Altmann et al., 2010; Chehab et al., 2022). Chehab et al. (2022) specifically have shown the power of PFI for analysing how certain language features like word frequency affect the performance of forecasting MEG data at different temporal and spatial locations. Csaky et al. (2023) have validated the use of PFI in MEG decoding against more traditional methods, such as sliding-window decoding. We leverage PFI in this paper due to its flexibility in obtaining spatio-temporal information from trained models, and its applicability to non-linear models. This method allows for direct assessment of both spatial and temporal information by permuting across time points (for each channel) and across channels (for each time point), respectively. We call these temporal and spatial PFI, respectively. Specifically, for temporal PFI we take a trial of dimension channels × timesteps and select a window of 100 ms around each timestep. Within this window the channels are shuffled to create random data, while the rest of the data are left untouched. By computing the accuracy on all trials with the respective windows disrupted we obtain an accuracy timecourse. The greater the decrease from the original accuracy (with unpermuted inputs) the more in (visual) stimulus-related information is present in that time period. Similarly for assessing spatial information content we take each channel (or group of nearby channels) and shuffle the timesteps within this channel to create random data, while the rest of the channels are left untouched. The greater the decrease from the original accuracy the more stimulus-related information is present in that channel, and we can repeat this for all channels to obtain sensor space maps.

---

[2]Since we do not use padding, this excludes the amount that gets chopped off because of the kernel size itself.

We also extended the PFI method to individual kernels of the Wavenet model. In this case, the feature importance measure is the absolute difference between the kernel output using the original and permuted inputs. We reason that a more important feature will cause a higher output deviation. The model receives the same permuted inputs as in model-level PFI (described previously), the difference is that we look at the output of individual kernels instead of the whole model. Spectral PFI, introduced in Csaky et al. (2023) is also run at the kernel level. This method can assess which frequency bands contribute the most to the observed performance, or in our case which frequency bands are kernels most sensitive to. The method works by first computing the Fourier transform of the trial, then shuffling the Fourier coefficients within a specific frequency range (for all channels), and computing the inverse Fourier transform to go back to the time domain. The greater the accuracy decrease of such a disrupted trial from the original accuracy the more stimulus-related information that frequency band carries. By repeating this process for all frequency bands we obtain an accuracy loss plot with respect to frequency. In this paper we only apply spectral PFI at the kernel level as previously described, thus assessing the frequency-sensitivity of specific kernels within the model.

## 2.4 | Experimental details

Our main evaluation metric is the classification accuracy of the across-subject decoding across the 118 classes. Recall that in across-subject decoding, each subject has a train and test split, and the aim is to see if a single group decoding model generalises across (and within) subjects. Train and validation splits with a 4:1 ratio were constructed for each subject and class. This means that classes are balanced (i.e., contain the same number of examples) across subjects and splits. Subject-level and group-level models are trained and evaluated on the same splits. Note that for each model, an extra training is conducted wherein the (linear) identity function is used as an activation function to assess the influence of non-linearity. Linear and non-linear models are trained for 500 and 2000 epochs (full passes of the training data), respectively, with the Adam optimiser (Kingma & Ba, 2015). Table 2 lists all of the model and training combinations that are presented in Figure 3. In this section when we refer to *model* we mean specific trained models on the respective datasets given in Table 2.

Dropout was set to 0.4 and 0.7, and a batch size of 590 and 59 was used for group-level and subject-level models, respectively. The learning rate was set to 0.0001 for group-level, and 0.00005 for subject-level models. Training of a single subject-level and group-level model took 5–15 min and 4 h on an NVIDIA A100 GPU, respectively. For linear models, validation losses (cross-entropy) and accuracies were negatively correlated, that is, loss decreases while accuracy increases, and eventually both suggested overfitting. Since non-linear models are more expressive, they overfitted sooner according to the loss, but accuracy kept improving until it reached a plateau, never overfitting. Analysing the loss distribution across validation examples (for non-linear models) shows that even during overfitting most examples' loss keeps decreasing with a few high-loss outliers disproportionately influencing the mean. Since accuracy is binary, outliers are diminished, explaining the apparent difference in learning behaviour. For linear models, this unintuitive behaviour was not observed probably due to inherent model simplicity.

We compute Wilcoxon signed-rank tests for comparisons of interest over trained models, where the pairing is within-subject, and samples are the subject-level mean accuracies over validation trials. We used PyTorch for training (Paszke et al., 2019) and several other packages for analysis and visualisation (Pedregosa et al., 2011; Virtanen et al., 2020; Harris et al., 2020; Wes McKinney, 2010; Waskom, 2021; Hunter, 2007).

## 3 | RESULTS

### 3.1 | Subject embedding closes the gap between subject-specific and group models

Validation accuracies for all trained models are shown in Figure 3. Interestingly, at the subject level, linear models performed slightly better than non-linear (4% increase, $p = 5.7e-4$). We think that both the limit in data size and noise levels in the data contribute to the subpar performance of non-linear models when trained/validated within-subject. The large between-subject variability common to MEG datasets is apparent, with individual subjects' accuracy ranging from 5% to 88%. As expected, training naive group models, that is, a naive application of either the linear or non-linear WaveNet Classifier to the group modelling problem (orange violin plots), results in much worse performance than training subject-level models, that is, 30% decrease compared to linear subject. Inferring such high variability implicitly between so few subjects is not trivial. Adding subject embeddings to the non-linear model (non-linear group-emb) improves performance by 24% ($p = 1.9e-6$), with no increase for the linear model (linear group-emb). This shows that leveraging subject embeddings in conjunction with non-linear activations can narrow the gap with subject-level models (6% difference with linear subject, $p = 1.3e-2$). Limiting the non-linearity to the first layer resulted in a subpar performance, similar to that of a linear model. This indicates that non-linearity is needed within multiple layers to benefit from subject embeddings. The impact of subject embeddings is further investigated in Section 3.2.

TABLE 2  Model and training combinations and their corresponding naming.

| Method name | Linear/non-linear | Number of conv layers | Subject embeddings | Trained on each *subject* or on concatenated *group* data | Further finetuned on individual subjects |
|---|---|---|---|---|---|
| Linear subject | Linear | 3 | No | Subject | No |
| Non-linear subject | Non-linear | 3 | No | Subject | No |
| Linear group | Linear | 6 | No | Group | No |
| Non-linear group | Non-Linear | 6 | No | Group | No |
| Linear group-emb | Linear | 6 | Yes | Group | No |
| Non-linear group-emb | Non-linear | 6 | Yes | Group | No |
| Non-linear group finetuned | Non-linear | 6 | No | Group | Yes |
| Non-linear group-emb finetuned | Non-linear | 6 | Yes | Group | Yes |

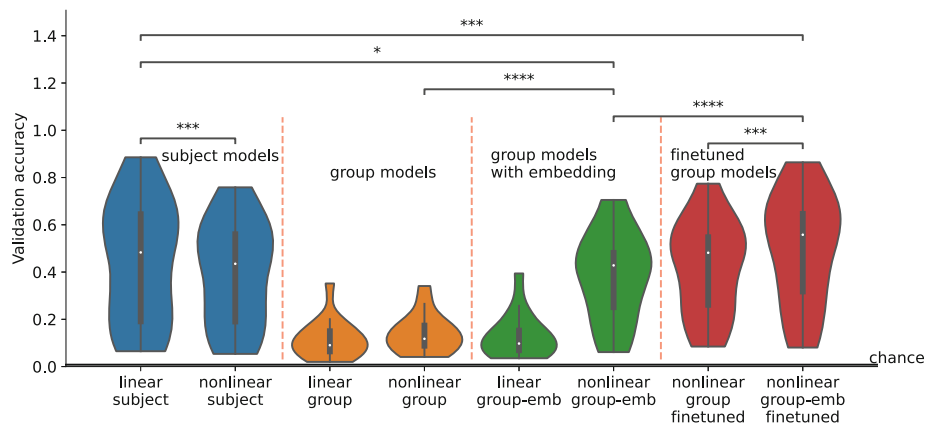

FIGURE 3  Trained subject-level and group-level models evaluated on the validation set of each subject. Wilcoxon signed-rank tests are shown for comparisons of interest ($* = p < 5e-2, ** = p < 1e-2, *** = p < 1e-3, **** = p < 1e-4$). The non-linear group-emb finetuned model is finetuned separately on each subject, initialized with the non-linear group-emb model. Chance level is 1/118.

We also finetuned the embedding-aided group-level model on the training data of each subject separately (non-linear group-emb finetuned) for 500 epochs. We effectively use the group-level model as an initialisation for subject-level models, improving over subject-level models trained from scratch (linear subject), achieving 50% accuracy (5% increase, $p = 1e-3$). This shows that representations learned at the group level are useful for subject-level modelling. In contrast, finetuning a naive group model (non-linear group finetuned) only achieved 42% accuracy showing that the best performance is reached when finetuning is combined with the best group-model. Thus, in addition to closing the gap between subject-level and group-level modelling, finetuning our embedding-aided model provides the best overall accuracy for subject-level modelling. The variance of non-linear group-emb (19%) and non-linear group-emb finetuned (24%) is lower than linear subject (26%). Generally, group models reduce between-subject variability.

In neural decoding, group models are widely understood to perform worse than individual models (Dash et al., 2020a; Guggenmos et al., 2018) But why is this? By plotting per-subject performance in both kinds of models (Figure 4), we see something revealing. In the case of non-linear group-emb, 4 subjects with generally low accuracies (15%–30%) had higher accuracies than linear subject (even though the mean across subjects is lower). This suggests that group models could be successfully used for some subjects if those subjects could be identified. Indeed, strong negative correlations of −0.88 and −0.54 are obtained between linear subject subject-level accuracies and the change in accuracy achieved by the non-linear group-emb and non-linear group-emb finetuned models, respectively. Comparing finetuning to from-scratch subject-level models (linear subject), only 2 high-accuracy subjects are slightly worse and generally low/mid-accuracy subjects show more improvement than high-accuracy subjects (Figure 4).

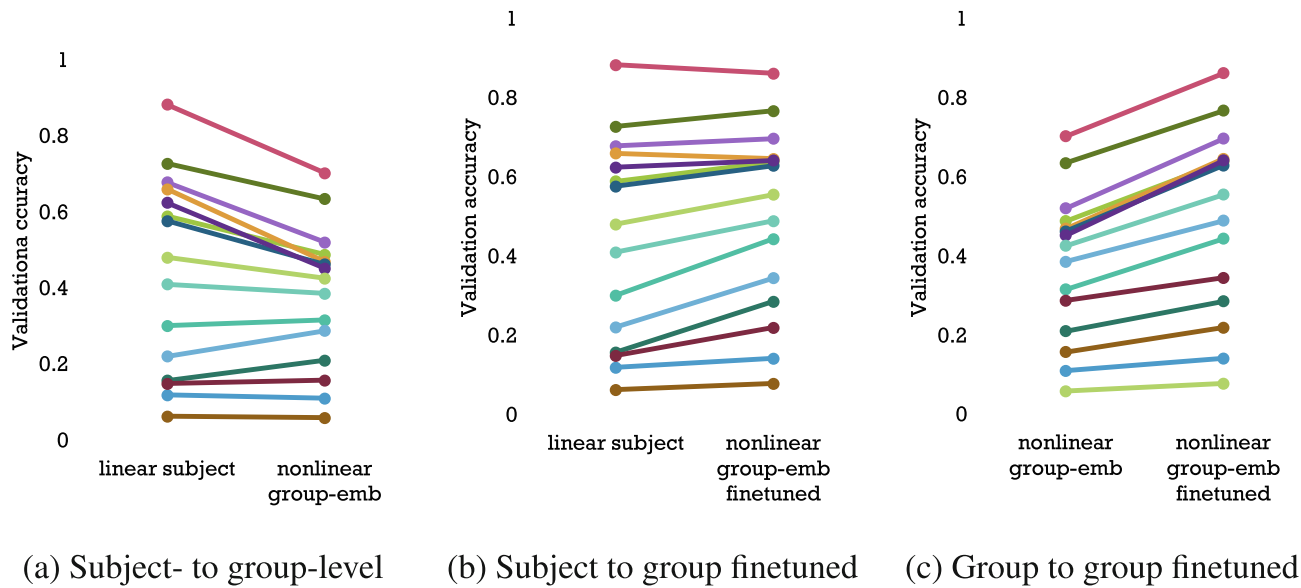

**FIGURE 4** Accuracy changes across all 15 subjects (individual colours), when comparing trained linear subject, non-linear group-emb and non-linear group-emb finetuned models. Both non-linear group-emb and the finetuned version clearly reduce the variability of accuracies across subjects and are especially helpful for low-accuracy subjects. When finetuning non-linear group-emb on individual subjects (c), we can see that accuracy increases for all subjects, and especially for high-accuracy subjects. This is unsurprising because these subjects have good enough data on their own for subject-level models to be able to learn well. As seen in (a) and (b) these high-accuracy subjects are usually impaired by group-level models, exactly for the aforementioned reason.

In addition to the main dataset used in this section, we also analysed our main findings on another publicly available visual MEG dataset and found results with a similar pattern (see the Supplementary Material).

In summary, these results suggest the following recommendations for decoding MEG task data. (1) Subject embeddings and non-linearity should be used for achieving good group models. (2) Group-level models can be used to improve over subject-level models on low-performance subjects. (3) For the best subject-level performance, the finetuning approach should be used, benefitting low-performance subjects the most.

## 3.2 | Insights into subject embeddings and other modelling choices

For the embedding-aided group-level setup (non-linear group-emb), 4 further models were trained for 5-fold cross-validation. The training and validation sets still contained 80% and 20% of trials respectively, and the splitting was done so that all of the data appears exactly once in the validation set across the 5 folds, for each subject and class. Average accuracy was 37.4% (as opposed to the 38% reported in Figure 3), with a 95% confidence interval of 0.8%. Thus, the proposed group-level model is robust to different random seeds and dataset partitions. More extensive robustness analysis is omitted due to computational constraints.

In non-linear subject-level models (non-linear subject), accuracy improves as we use fewer convolutional layers, whereas for non-linear group-level models (non-linear group-emb) using more layers improved accuracy (see Table 3) Thus, subject-level models seem to rely more on the fully-connected block as they are unable to extract good features, and group-level models rely more on the convolutional block to learn shared features across subjects. To have a fair comparison between the two approaches we selected the best number of layers for both individual and group-level models. To be clear, because of how we perform the temporal downsampling after the convolutional layers (described in Section 3.1), using fewer convolutional layers increases the overall parameter count because the fully-connected block has to be enlarged. Thus, the group model (with 6 conv layers), is about 2.5× smaller than the subject-level models (with 3 conv layers). However, non-linear group-emb finetuned models achieve higher accuracy than from-scratch subject-level models linear subject. This shows that, when they are initialised well (with a group model trained on multiple subjects), even subject-level models can benefit from non-linearity and more convolutional layers.

We tried different approaches to understand how subject embeddings help the non-linear group-emb model. A clustering or 2D projection of the embedding space such as PCA or t-SNE (van der Maaten & Hinton, 2008) did not show any clusters (see Figure 4 in the Supplementary Material). This is likely to be a consequence of only having 15 subjects since cases where such visualisations work well (Liu et al., 2017) typically have thousands of dimensions (e.g., words in word-embeddings). To assess whether the embeddings simply encode which subjects are good, we transformed the embeddings with PCA

| | Linear subject | Non-linear subject | Non-linear group-emb |
|---|---|---|---|
| 3 conv layers | 0.45 | 0.39 | 0.22 |
| 6 conv layers | 0.41 | 0.25 | 0.38 |

**TABLE 3** Effect of number of convolutional layers on the validation accuracy of two subject-level and one group-level model.

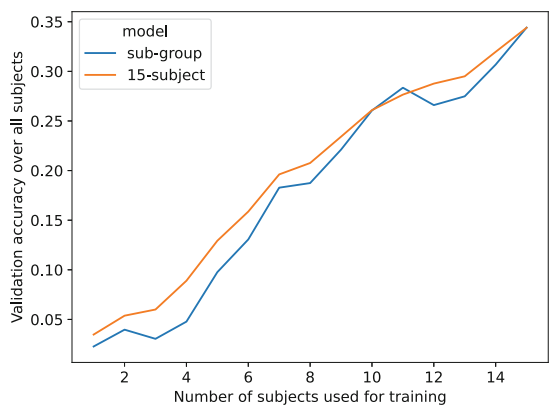
(a) Evaluation over all subjects

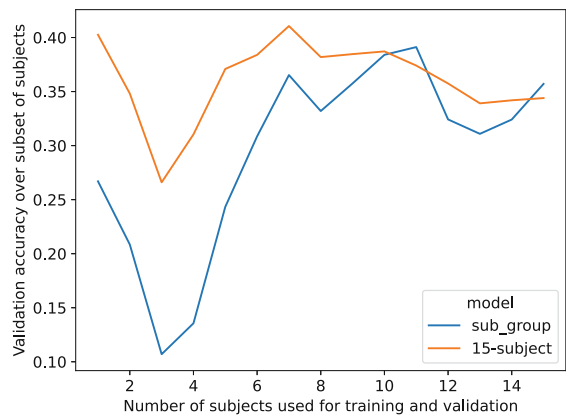
(b) Evaluation over subsets of subjects

**FIGURE 5** (a) Validation accuracy over all subjects with respect to increasing the subset of subjects used for training the sub-group model (blue line) on the horizontal axis. The 15-subject model (orange line) is our standard non-linear group-emb model trained on all subjects. (b) Validation accuracy over the subset of subjects used for training the sub-group model (blue line). The 15-subject model (orange line) is our standard non-linear group-emb model trained on all subjects. The 15-subject model is evaluated on the same increasing sets of subjects as used for the sub-group models.

and correlated all components with the accuracies across subjects. We found no significant correlations; therefore, embeddings do not appear to encode information about subject-level accuracy. To assess how much embeddings contribute to a trained model, we tried both setting the embeddings to zero and shuffling them. The validation accuracy of non-linear group-emb decreased to 10% for both approaches. This is a 28% reduction from the original accuracy. Thus, embeddings encode crucial information to aid decoding, but the non-linear group-emb is still better than chance without them. For further insights see the Supplementary Material.

Training with an embedding dimensionality of 3 and 14, resulted in 20% and 38% accuracy, respectively. We tried these two settings to see how embedding size in the lower and upper limits influences performance. As an embedding dimensionality of 14 performs the same as 10, we could draw the conclusion that 10 is not a limiting factor. From the much worse result with an embedding dimensionality of 3, we could draw the conclusion that compressing the embedding representations too much is not possible. As with the clustering analysis, this is likely to be due to having few subjects.

To see what effect increasing the number of subjects has on group model performance we trained 15 embedding-aided sub-group models with increasing number of subjects, that is, 1 subject, 2 subjects, …, 15 subjects. We used the same hyperparameters as for our original non-linear group-emb. We then evaluated each sub-group model on the validation set of the subjects it was trained on (Figure 5b). The resulting validation accuracy is shown in the plot as a function of the number of subjects in the sub-group. One downside of this approach is that the order in which we add the subjects to the sub-group models matters a lot because of the high between-subject variability. However, to test multiple orderings we would have to run hundreds of trainings which is not possible under our computational constraints. Nevertheless, we compared our increasing subject-number sub-group models with the theoretical best performance achieved by the group model trained on all subjects (15-subject). We can see that the gap between the full group model and the restricted sub-group models generally tightens as we increase the number of subjects used for training. It is difficult to draw strong conclusions without repeating this analysis with different permutations of subjects.

An alternative visualization for the previous analysis is to keep the validation set fixed, that is, always compute validation performance on the validation set of all subjects (Figure 5a). To provide the theoretical maximum from the 15-subject group model we took its performance on the respective subjects (e.g., in the case of 2 subjects, the first 2 subjects), and replaced the other subjects with a 1/118 (chance) accuracy value. This again shows a slight tightening between the full group model and the restricted sub-group models as we increase the number of subjects. Notably, there is a dip in performance when we add subject 12 to the group model as this subject had a particularly bad performance.

## 3.3 | Leave-one-subject-out evaluation

To this point, we have reported results for across-subject decoding, in which we use a single group decoding model that generalises across

(and within) subjects; an approach that is, for example, relevant when one wants to gain neuroscientific insights that generalise to the group level. In this section, we report leave-one-subject-out (LOSO) cross-validation results; which is relevant, for example, when one wants to develop BCI methods that work on previously unseen subjects. Movement classification is one such application where it would be beneficial to be able to use a decoder trained on other subjects in a zero-shot setting. We also analyse how performance improves when we allow models to use increasing amounts of data (finetuning) from the left-out subject. We compare the LOSO and finetuning performance of non-linear group, non-linear group-emb and linear subject. The linear subject approach serves as a baseline and it is only trained on the left-out subject. Thus, in the LOSO (zero-shot) setting, this model has chance level, since no training is performed on the left-out subject.

When training non-linear group-emb the left-out subject's embedding was initialised randomly. In the LOSO (zero-shot) evaluation, both group models achieve 5% accuracy (Figure 6a). Up to the case when 70% of the training data of the left-out subject is used, both group models are much better than linear subject ($p < .05$, corrected for multiple comparisons). This is expected and the benefit of group-level models in LOSO analysis has been previously established (Elango et al., 2017). Thus, to achieve the same level of performance as linear subject much less data are needed when finetuning a group model. Unsurprisingly, the non-linear group-emb model does not improve over the naive model (non-linear group), but is, importantly, not worse. As opposed to the finetuning setup in Figure 3, when adapting to new subjects, better group performance does not translate to better finetuning performance. We think this is because when adapting to a new subject, that subject's embedding was randomly initialised, and thus it has to be learned during the finetuning. This is a limitation of our approach.

### 3.4 | Neuroscientific insights are gained from a deep learning based group-level model

An established critique of deep learning models applied to neuroimaging data are the lack of interpretable, neuroscientific insight they provide about the underlying neural processes that drive the decoding (Murdoch et al., 2019). To gain such insights, it is useful to assess the time- and space-resolved information/discriminability within trials. Figure 6b shows the temporal and spatial PFI of the trained non-linear group-emb model. To make the results robust and smooth, the shuffling for temporal PFI was applied to 100 ms windows, and magnetometers and gradiometers in the same location were shuffled together for spatial PFI. Time windows or channels with higher accuracy loss than others are interpreted as containing more information about the neural discriminability of the visual images. This indicates when and where information processing related to the presented images is happening in the brain.

Temporal PFI shows a large peak around 150 ms which is in line with previous subject-level PFI results on this dataset (Csaky et al., 2023). After this, the information content rapidly decreases, with a second, smaller peak around 650 ms, which could correspond to a brain response following the end of image presentation at 500 ms. Spatial PFI shows that the most important channels are in the back of the head in the sensors in visual areas, as expected for a visual

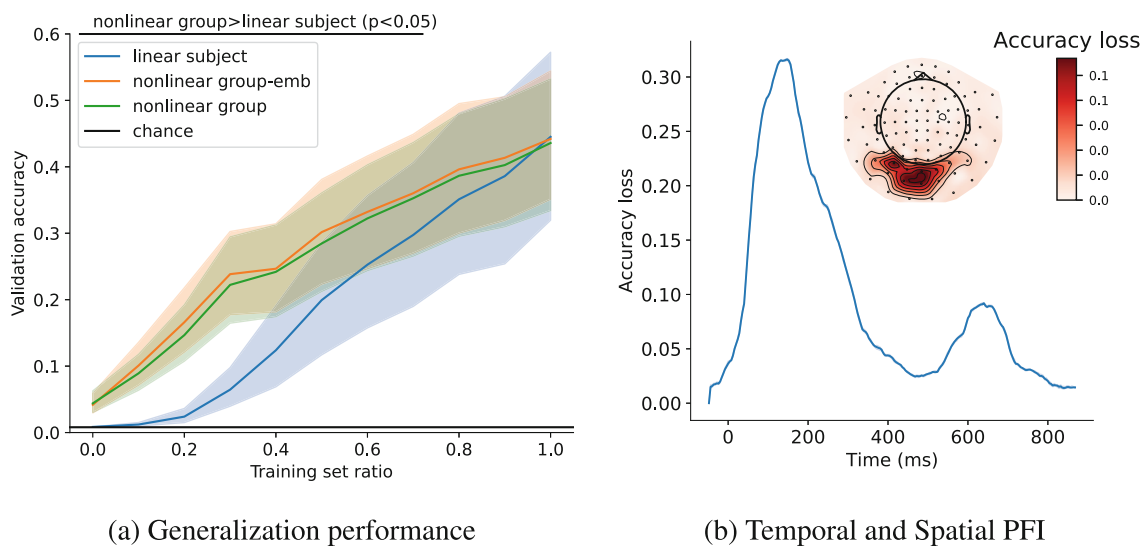

(a) Generalization performance  (b) Temporal and Spatial PFI

**FIGURE 6** (a) Generalisation and finetuning on left-out subjects. The horizontal axis shows the amount of training data used from the left-out subject; a training set ratio of 0 corresponds to a zero-shot approach. Linear subject is trained from scratch, while non-linear group-emb and non-linear group are initialised with the trained non-linear group-level model with and without embeddings, respectively. The 95% confidence interval of the accuracy across left-out subjects is shown with shading. (b) Temporal (line) and spatial (sensor space map) PFI for the trained non-linear group-emb model. For temporal PFI accuracy loss (vertical axis) is plotted with respect to time since visual image presentation (horizontal axis). Shading shows the 95% confidence interval which is not visible due to low variability. For spatial PFI, darker red shading is equivalent to higher accuracy loss.

task. We found good agreement between this PFI analysis and the alternative approach of a gradient-based analysis often used in deep learning models (see Figure 2 in the Supplementary Material).

## 3.5 | Weights encode meaningful spatio-temporal and spectral information

To provide further insight into our trained non-linear group-emb model, we next show that interpretable spatial, temporal and spectral information can be obtained by analysing the learnt weights. This analysis becomes possible because we use a multi-layered neural network, and there is no equivalent analysis that we could do in a classical linear model. When using deep learning it is important to ask how the trained model arrives at the information presented in Section 3.4. We can leverage the structure of the model, that is, the successive layers, and the filters in the convolutional layers can be regarded as individual computational units. The aim here is to understand the model itself and how it represents and processes the data internally. This is in line with previous efforts showing how successive layers in a deep convolutional model align with the visual system of the brain (Kriegeskorte, 2015).

Figure 7 shows results for just 3 of the 6 convolutional layers, with all 6 layers shown in Figure 5 and Figure 6 in the Supplementary Material. Kernels within a layer tend to have similar temporal sensitivity, and hence we only show 5 out of over 1e5 total kernels (Figure 7c). Output deviations are standardised to compare temporal PFI across kernels with different output magnitudes. In the early layers, sensitivity peaks around 100 ms (as in Figure 6b), then rapidly decreases, eventually climbing again slowly. Kernels in early layers have somewhat random spatial sensitivity (Figure 7a), but this gets narrowed down to channels over the visual cortex in deeper layers, with some differences between individual kernels. This sensitivity is similar to the spatial features that were shown to be most informative for classification performance (see Figure 6b).

Figure 7b shows the temporal profile of the spatial PFI. This is achieved by limiting the shuffling to 100 ms time windows and 4-channel neighbourhoods (3 closest channels for each channel) at a time, which is then repeated across all time points and channels. This shows spatial sensitivity does not seem to change with time; that is, the most important channels are always the same, also observed in previous spatiotemporal PFI analyses of this dataset (Csaky et al., 2023).

In neurophysiology, we are often interested in the oscillatory content of the signal, and what/how specific frequencies are associated with certain tasks, here, decoding of visual stimuli. To this end, we use PFI in the spectral domain, where it is used to measure the change in kernel output to perturbations in specific frequency bands (Figure 8a). Please see Figure 7 in the Supplementary Material for all 6 layers. Across all layers and kernels, the profile has a 1/f (frequency) shape with a clear peak at 10 Hz. These are common features of the MEG signal (Demanuele et al., 2007; Drewes et al., 2022), indicating that the spectral sensitivity of the kernels coincides with the power spectra of the data. Interestingly, a previous model-level spectral PFI analysis on this dataset did not show the 10 Hz peak (Csaky et al., 2023). Thus analysing the weights of deep learning models can reveal more or different information. In Figure 8b, we also looked at the spectral PFI of 4-channel neighbourhoods and found that kernels are sensitive to the same channels (in the visual area) across all frequencies, with these channels having larger 10 Hz peaks.

Kernel FIR analysis shows the power spectra of kernels' outputs when input examples are Gaussian noise (Figure 3 in the

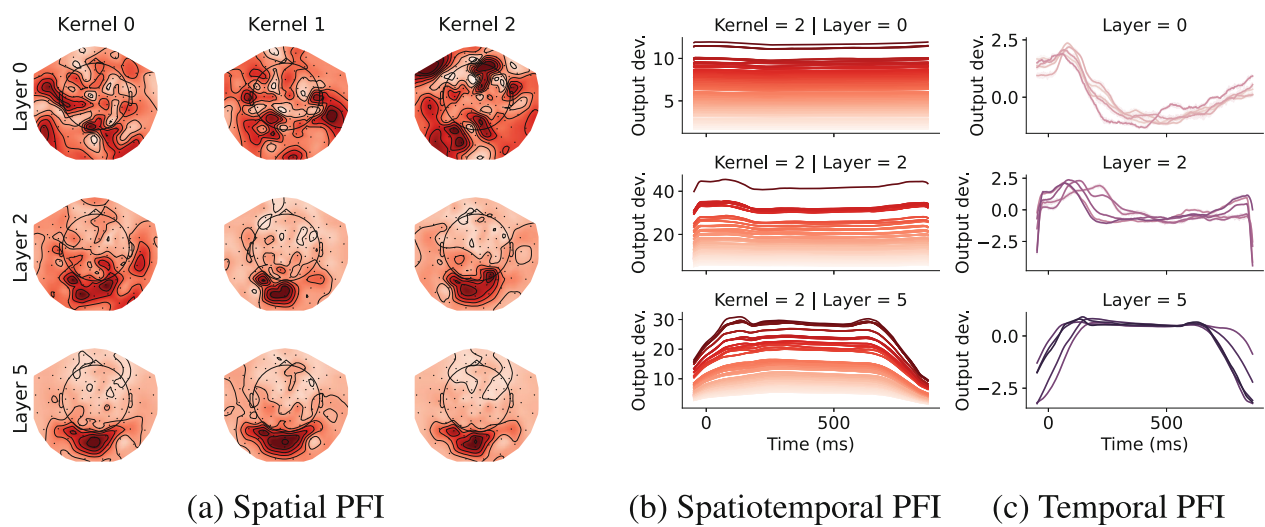

(a) Spatial PFI   (b) Spatiotemporal PFI   (c) Temporal PFI

**FIGURE 7** Spatio-temporal insights can be obtained using PFI. Spatial (a), channel-wise temporal (b) and temporal (c) PFI across non-linear group-emb kernels within 3 layers (rows). For spatial PFI, kernels are plotted separately; whereas for temporal PFI, 5 kernels (lines) are plotted together. Channel-wise temporal PFI shows the temporal PFI of each channel for Kernel 2. Channel colouring is matched to the corresponding spatial PFI map, and darker reds mean higher output deviation. For temporal PFI, output deviation is normalised. The horizontal axis shows the time elapsed since the image presentation, for both temporal PFI types. 95% confidence intervals are shown with shading.

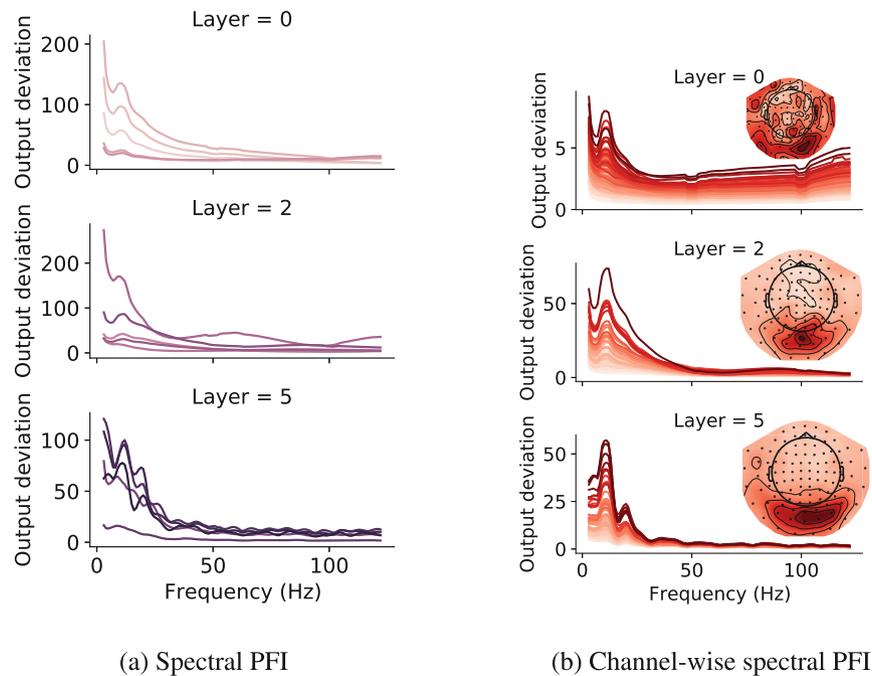

FIGURE 8 Frequency sensitivity of kernels via spectral PFI (a), channel-wise spectral PFI (b), and frequency characteristics via kernel FIR analysis (c), from 3 layers (rows). Kernels are plotted together (lines) for spectral PFI, and in separate columns for kernel FIR analysis (normalised). Each channel-wise spectral PFI plot is for 1 kernel, where lines show the spectral PFI of corresponding channels in the sensor space map. 95% confidence intervals are shown with shading for spectral PFI. Due to small variability across permutations, this is barely visible. For spectral PFI the band-width was set to 5 Hz to obtain a smooth frequency profile.

Supplementary Material). Both the spectral PFI and kernel FIR analysis show that there is significant variability between the spectral information encoded by various kernels.

In summary, the analyses presented in this section show that the kernels are sensitive to interpretable temporal, spatial and spectral features of the MEG data. Specifically, we have shown that kernels are sensitive to channels in the visual area, with this sensitivity getting more focused in deeper layers. Kernels are also sensitive to the 10 Hz peak of the MEG data, and the temporal sensitivity shows a peak at 100–150 ms in early layers.

## 4 | DISCUSSION

In this work, we focused on across-subject decoding, motivated by the fact that group-level models that perform well in this manner can be useful for gaining neuroscientific insights that are relevant at the group level. In this setting, our proposed deep learning-based group-level model outperforms naïve group models and achieves similar performance to subject-level models, but with three key benefits. First, it provides potentially richer insights at the group level. As non-linear models are required for the group-level decoding to work, we had to use PFI, and showed that it is effective in the case of non-linear group-level models. Second, there is potential for the group-level model to outperform subject-level models when larger population datasets are available. Third, a group-level model can be used to initialise subject-level models surpassing the performance of subject-level models initialised randomly. We have shown how subject embeddings and non-linearity are crucial for this. These are important insights towards the goal of using group models in decoding neuroimaging data, which would allow for better use of this inherently limited resource.

Interestingly, we found that at the subject level, linear models perform better than their non-linear counterparts. Although some studies have found deep learning to improve over simpler linear models, this improvement is often marginal (Cooney, Korik, et al., 2019; Schirrmeister et al., 2017). Such results are difficult to generalise across different MEG datasets, due to variability in tasks, the number of subjects, and the amount and quality of data (Schirrmeister et al., 2017).

Other than being useful for fine-tuning, our embedding-aided group model can be useful in the case of much larger datasets, where we cannot afford to have a separate model for each subject. As we have shown, even in this limited dataset with 15 subjects, the group model can provide improvement in a few subjects. Our results suggest follow-up studies to understand why some subjects performed better or worse.

As mentioned in Section 2.3, permutation feature importance (PFI) is a suitable, model-agnostic measure for assessing the time- and space-resolved information within trials. More traditional methods for gaining such insights are for example sliding window decoding and spatial spotlight decoders. Sliding window decoding can provide group-level temporal information by averaging the predictions over individual subject models (Csaky et al., 2023). As we have shown non-linear models trained on the full epoch are required for effective group-level modelling, prohibiting the use of sliding-window or spatial spotlight decoding. PFI can also provide group-level temporal, spatial and spectral information by averaging over linear subject-level models (Csaky et al., 2023). Here our aim was to show that PFI works similarly well in the case of non-linear group-level models.

We have demonstrated the use of PFI on group models to obtain insight into which time points and channels contributed to the decoding and to obtain meaningful information encoded in convolutional kernels. Using this and other methods, such as representational similarity analysis, neuroscientific investigations can be performed at the group level using a single model, instead of averaging over individual subject models. We note that one downside of PFI is that the absence of influence on the output does not necessarily mean that a specific channel or time window does not carry information about the target variable. However, empirically, PFI provides very similar results to standard methods (Csaky et al., 2023) on the same data that we used in this work.

While the across-subject decoding we focus on in this work is most relevant to situations where we want to obtain insights at the group-level, other applications, such as BCIs that need to work well on previously unseen subjects, may be more appropriately evaluated using leave-one-subject-out (LOSO) evaluation. In this context, we found that using subject embeddings did not improve performance. Exploiting subject embeddings in a pure LOSO framework is not trivial, as some additional approach is needed to initialise/learn the embedding of the left-out subject in an unbiased manner. We have not tried to only optimize the subject embedding while freezing the rest of the model. While computationally less expensive, this is not expected to be as good as optimizing the whole model (and subject embedding) on the new subject, which we have presented in Section 3.3. In larger datasets with more subjects, between-subject similarities in the embeddings could be exploited and different heuristics explored, for example, initialising the embedding with the average of all learned subject embeddings. However, research aimed at improving performance in new subjects often leverages transfer learning in some way, where a limited amount of data from the new subject can be used (Zubarev et al., 2019). In this scenario, we think our across-subject group model could be helpful, by, for example, using the limited data from the new subject or by learning a useful embedding for the new subject in an unsupervised manner. As we have shown in Section 3.1 this could be especially useful for subjects with low performance. As opposed to a naive continuation of the trends in Figure 6a, we expect that with more trials, the gap between group initialisation and training from scratch would continue, up to some limit. We believe that the reason why the gap closes at 100% training data are due to the ratio of training and validation sets and the low number of examples. The small validation set (6 examples per class) is probably not representative of the full data distribution.

Decoding can be applied to most tasks/modalities, such as images (Cichy et al., 2016), phonemes (Mugler et al., 2014), words (Cooney, Korik, et al., 2019; Hultén et al., 2021), sentences (Dash et al., 2020a), and motor movements such as imagined handwriting (Willett et al., 2021), jaw movements (Dash et al., 2020b), or finger movements (Elango et al., 2017). Here, we used image categorisation because it is a widely studied decoding task and we had access to a dataset which is relatively large for the field of neuroimaging. However, we expect our results to readily generalise to other decoding modalities.

Recently, different transfer learning approaches have been proposed to deal with the problem of variability between subjects. Kostas and Rudzicz (2020) have proposed two distinct methods. First, there is Euclidean alignment, which is very similar to a spatial whitening of the data. We tried this in conjunction with our group model, and found it to lower performance, and thus opted for a simpler channel-wise standardisation. Second, there is mixup regularisation, which is entirely complementary to our approach and can be used in conjunction with it. It is a general regularisation/data augmentation technique and does not specifically deal with inter-subject variability. Most transfer learning frameworks consist of applying a model trained on one subject to a different (target) subject (Cooney, Folli, & Coyle, 2019; Dash et al., 2019; Elango et al., 2017; Halme & Parkkonen, 2018; Li et al., 2021; Olivetti et al., 2014). Some approaches use learnable affine transformations between subjects (Elango et al., 2017), while others finetune the whole model on target subjects (Cooney, Folli, & Coyle, 2019; Dash et al., 2019).

Hyperalignment has been successful for fMRI data to align different subjects to a common cortical space, and some applications have been explored in MEG data as well. For example, Benz (2020) used hyperalignment on MEG data via procrustes matrix transformation to a common sensor space and showed improvement in evoked fields. Similar methods have been explored in recent studies aiming to deal with between-subject variability (Michalke et al., 2023; Ravishankar et al., 2021; Zhou et al., 2020). However, to our knowledge, no prior work has applied it successfully to MEG decoding. One key consideration is that hyperalignment is a linear method, constraining the transformation between subjects. While this is a sensible assumption, we think that in order to fully leverage data from multiple subjects a non-linear method is required. Our subject embedding method is fully data-driven without any constraints on the nature of variability between subjects. It may be that this flexibility becomes truly useful when dealing with a large number of subjects, and for a few subjects, the linear assumptions of hyperalignment could work better. Our method also directly optimises the subject embeddings for the decoding objective. It is not clear, whether an unsupervised method, such as hyperalignment would result in better decoding accuracy. We leave it for future work to provide a full comparison between hyperalignment and subject embedding for MEG decoding.

Transfer learning is also popular in the wider machine learning field. Parallels can be drawn with domain adaptation (Long et al., 2015), or transferring knowledge from large to small datasets within the same domain (Wang et al., 2019, Zhuang et al., 2020]. Natural language processing (NLP) datasets often contain data from widely different sources (Radford et al., 2022), but due to the sheer size of the dataset and model complexity, training on joint data achieves good results (Brown et al., 2020; Devlin et al., 2019). Modern approaches to NLP often use Transformer-based (Vaswani et al., 2017) architectures. We believe convolutional architectures are an attractive approach for multi-channel timeseries data, and it remains to be seen how well Transformers would perform (Kostas et al., 2021). There are several issues to overcome when applying Transformers to multi-channel timeseries, similar but perhaps even more challenging than their application to computer vision (Dosovitskiy et al., 2020; Parmar et al., 2018). In our work, we wanted to keep the model relatively simple, as our dataset size is also limited, and analyse the effect of the subject embedding.

As discussed before, a naive concatenation of subjects does not work well on small neuroimaging datasets. Perhaps the most relevant parallels can be drawn with dialogue and speech modelling work, where inter-speaker differences are modelled using speaker embeddings (Li et al., 2016; Mridha et al., 2021; Saito et al., 2019; Zhang et al., 2018). Chehab et al. (2022) have similarly found that subject embeddings provide a small but significant improvement in encoding MEG data from a language task. They used a combination of recurrent and convolutional neural networks for encoding MEG data. However, limited information is provided on how subject embedding helps, and their results cannot be directly generalised to MEG *decoding*. Our results expand this work to the task of decoding images from MEG data and provide additional insight into how deep learning and subject embeddings help group-level decoding models. In concurrent work, Défossez et al. (2022) have also shown the effectiveness of subject embeddings in group-level speech decoding. They have also compared it to subject-specific layers as a way of dealing with between-subject performance and found this latter approach slightly better. The advantage of subject embeddings is that they use less parameters to deal with the between-subject variability and the structure in the learned representations can be readily interpretable.

We expect the subject embedding and group modelling to generalise to different task and recording modalities (EEG, fMRI, etc.) because they face similar decoding challenges. The specific Wavenet-based model is readily generalisable to other electrophysiological data such as EEG and electrocorticography (ECoG), because of the same temporal dynamics they capture. Further research is needed into deep learning models capable of implicitly learning inter-subject variability. An important question is whether scaling up models on large datasets would achieve this goal.

## ACKNOWLEDGEMENTS

This research was supported by the NIHR Oxford Health Biomedical Research Centre. The views expressed are those of the author(s) and not necessarily those of the NIHR or the Department of Health and Social Care. Richard Csaky is supported by a Wellcome Centre Integrative Neuroimaging Studentship. Mats W. J. van Es 's research is supported by the Wellcome Trust (215573/Z/19/Z). OPJ is supported by the UK MRC (MR/X00757X/1). Mark Woolrich's research is supported by the Wellcome Trust (106183/Z/14/Z, 215573/Z/19/Z), the New Therapeutics in Alzheimer's Diseases (NTAD) study supported by UK MRC and the Dementia Platform UK (RG94383/RG89702) and the EU-project euSNN (MSCA-ITN H2020-860563). The Wellcome Centre for Integrative Neuroimaging is supported by core funding from the Wellcome Trust (203139/Z/16/Z).

## CONFLICT OF INTEREST STATEMENT

The authors report no conflict of interest.

## DATA AVAILABILITY STATEMENT

Data sharing is not applicable to this article as no new data were created or analyzed in this study.

## ORCID

*Richard Csaky*  https://orcid.org/0000-0002-0028-3982

## SUPPORTING INFORMATION

Additional supporting information can be found online in the Supporting Information section at the end of this article.